\documentclass[runningheads]{llncs}
\usepackage{graphicx}
\usepackage{rotating}
\PassOptionsToPackage{hyphens}{url}\usepackage{hyperref}

\begin{document}
\title{Feature visualization for convolutional neural network models trained on neuroimaging data}
%
\titlerunning{CNN feature visualization for neuroimaging data}
%
\author{Fabian Eitel\inst{1, 2}\orcidID{0000-0003-2630-9172} \and
Anna Melkonyan\inst{1, 2}\orcidID{0000-0003-4836-2468} 
\and
Kerstin Ritter\inst{1, 2}\orcidID{0000-0001-7115-0020}}
\authorrunning{F. Eitel et al.}
%
\institute{Charité – Universitätsmedizin Berlin (corporate member of Freie Universität Berlin, Humboldt-Universität zu Berlin, and Berlin Institute of Health), Department of Psychiatry and Psychotherapy, Berlin, Germany \and
Bernstein Center for Computational Neuroscience, Berlin, Germany
\email{kerstin.ritter@charite.de}\\
}
\maketitle              
\begin{abstract}
 A major prerequisite for the application of machine learning models in clinical decision making is trust and interpretability. Current explainability studies in the neuroimaging community have mostly focused on explaining individual decisions of trained models, e.g. obtained by a convolutional neural network (CNN). Using attribution methods such as layer-wise relevance propagation or SHAP heatmaps can be created that highlight which regions of an input are more relevant for the decision than others. While this allows the detection of potential data set biases and can be used as a guide for a human expert, it does not allow an understanding of the underlying principles the model has learned. In this study, we instead show, for the best of our knowledge, for the first time results using feature visualization of neuroimaging CNNs. Particularly, we have trained CNNs for different tasks including sex classification and artificial lesion classification based on structural magnetic resonance imaging (MRI) data. We have then iteratively generated images that maximally activate specific neurons, in order to visualize the patterns they respond to. To improve the visualizations we compared several regularization strategies. The resulting images reveal the learned concepts of the artificial lesions, including their shapes, but remain hard to interpret for abstract features in the sex classification task.

\keywords{neuroimaging  \and convolutional neural networks \and explainability \and feature visualization.}
\end{abstract}
\section{Introduction}

In order to support the adaption of machine learning (ML) into clinical practice several methods for model interpretation have been developed. The field of explainable AI (xAI) has proposed a multitude of attribution methods, which evaluate the relevance of each pixel. Attribution methods typically use some form of backpropagation in order to generate heatmaps. Widely used methods include layer-wise relevance propagation, DeepLIFT, PatternAttribution and SHAP \cite{montavon2019layer,DBLP:journals/corr/ShrikumarGK17,kindermans2017learning,lundberg2017unified}. In the field of neuroimaging, it was shown that the generated heatmaps overlap with known biomarkers such as atrophy in the limbic system in Alzheimer's disease \cite{10.3389/fnagi.2019.00194,dyrba2021improving} or white matter hyperintensities in multiple sclerosis \cite{EITEL2019102003}. While these methods are relevant for explaining model decisions, the heatmap \(y\) is always dependent on the input \(X\) and the model parameters \(\theta\): \(y = F(X, \theta).\) In order to study the actual concepts that individual neurons have learned, methods are required that are independent of any specified input \(y = F(\theta).\)

In feature visualization studies, individual (or collections of) neurons are studied by iteratively generating images that maximally activate them. The resulting images reveal information about the learned concepts as well as the hierarchy of the models. Similar to the human visual cortex, CNNs process information in layers. The lowest layers (closest to the input) tend to focus on general concepts such as edges and colors, whereas higher layers tend to focus on more abstract concepts such as eyes and ears. While these abstract concepts are easily explainable in natural images, it becomes a lot more challenging in medical images with heterogeneous disease patterns. In a few neurological disorders, radiologists might detect visual distinct patterns such as the hummingbird sign, a pattern of midbrain atrophy in progressive supranuclear palsy \cite{graber2009teaching}. However, in the most common ML applications in neuroimaging today, such as Alzheimer's disease detection, brain age estimation and lesion and tumor segmentation, differentiating patterns are substantially varying. Feature visualization could eventually aid in finding reliable visual patterns that could help to identify brain disorders more easily.

Using feature visualization through naive optimization of an image leads to noisy images that are not interpretable, similar to those of adversarial examples \cite{goodfellow2015explaining}. Several regularization methods for feature visualization have been proposed, which force the learning process to create more reliable images \cite{mahendran2015understanding,mordvintsev2015inceptionism,tyka2016bilateral}. To the best of our knowledge, the features of models trained in neuroimaging have not yet been studied using those methods. In this study, we have tested several regularization methods in order to create images that maximally activate a channel of CNNs trained on structural magnetic resonance imaging (MRI) data. As a result, we show that lower level patterns are quite similar to those found in natural image studies, which supports the idea of transfer learning from natural images to MRI data. Furthermore, abstract patterns on artificial data intuitively represent the created signal, whereas abstract patterns in classification of an actual MRI task (sex classification) remain ambiguous. 

\section{Materials and Methods}
\subsection{Data}
\begin{figure}
    \centering
    \includegraphics[width=0.7\linewidth]{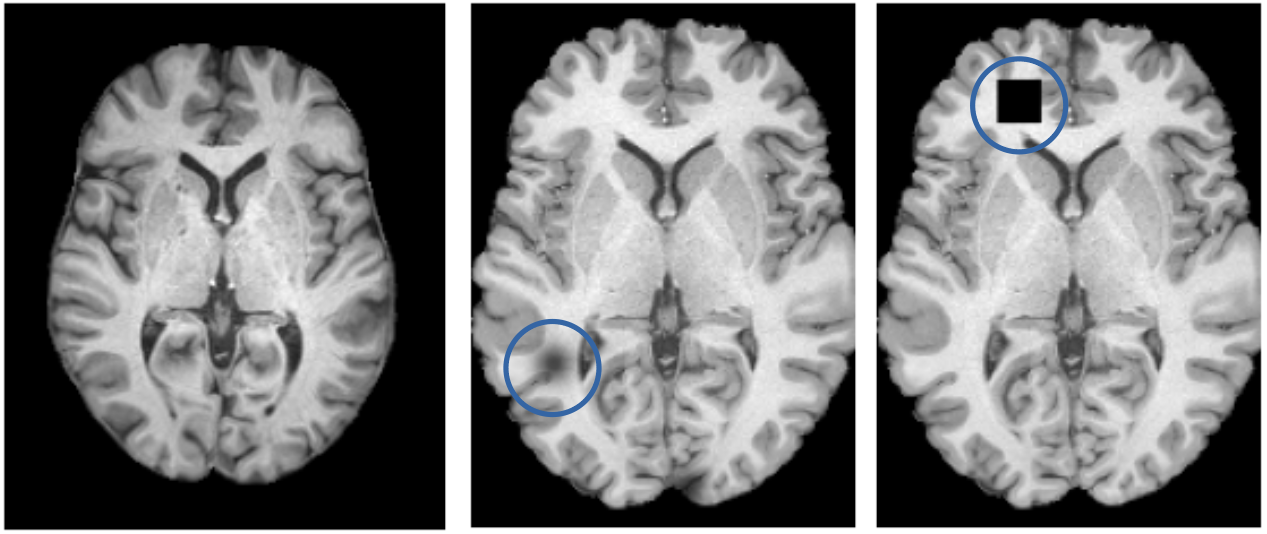}
    \caption{Example original MRI scan (left), and scans with artificial Gaussian (center) and square lesions (right). Lesions have been highlighted with circles.}
    \label{fig:brains}
\end{figure}

We have built three subsets based on the UK Biobank database\footnote{\url{https://www.ukbiobank.ac.uk/}}. The first data set consists of 3D MPRAGE sequences from 1,854 mostly healthy individuals (F: 1005, M: 849) and was used to train a sex classifier. The data comes from different sites and was selected randomly from the larger UK Biobank cohort. We acquired the pre-processed and pseudonymized versions of the data. It has been non-linearly registered to MNI152 space using FMRIB. For full pre-processing details see \cite{ALFAROALMAGRO2018400}. The full size (182, 218, 182) sequences were used, and each image has been further intensity normalized between 0 and 1. The other two data sets are based on 136,080 2D axial slices which have been selected from the first data set at random and have been modified using artificial lesions as shown in Figure \ref{fig:brains}. The second data set uses circular lesions with Gaussian blur to soften the edges, and the third data set uses square lesions with sharp edges. Here, the idea is to have one data set with striking, high frequency patterns (square lesions), one with more natural looking, albeit still synthetic patterns (Gaussian lesions). Approximately half of the slices were modified in both synthetic data sets and the model was trained to classify between lesioned slices and controls. The empty space around the brain in each 2D slice has been cropped yielding a final size of 140x192 pixels.

\subsection{CNN Model}
The model trained is a simple, VGG-net inspired 5-layer CNN \cite{simonyanzisserman2015}. After the first, second and fifth convolutional layer a max-pooling layer reduces the dimensionality. The convolutional layers all use 3x3x3 kernels and have 8, 16, 32, 64, and 64 filters in the given order, while max-pooling layers use 3x3x3 kernels and 4x4x4 kernels on the last layer. Furthermore, max-pooling layers use a stride of 1, 2, 2 in the given order. All convolutional layers are followed by a ReLU non-linearity. The model was initially 3D and then converted to 2D for the lesion classification tasks.

On the 3D sex classification task 20\% of the data were reserved for the test set, while from the remaining data another 20\% was separated as a validation set. The Adam optimizer was used with a learning rate of 0.0001 and weight decay of 0.00001. Early stopping was used with a patience of 10, and batch size was set to 8.
Gaussian and square lesion models were trained on the respective data sets using the Adam optimizer with a learning rate of 0.001 and weight decay of 0.0001. The data sets were split to include 81,135 training, 27,135 validation and 27,810 testing images. On both data sets we trained in batches of 200 images, and early stopping with a patience of 7 iterations. The models achieved good predictive performance for sex classification (92.22\% balanced accuracy) and almost perfect performance on Gaussian lesion classification (99.61\% bal. acc.) and square lesion classification (99.35\% bal. acc.).

\subsection{Feature Visualization}
Feature visualization is an explanation method, which has been developed to investigate the inner workings of CNNs. In contrast to image-dependent heatmaps which show \textit{attribution}, feature visualization methods generate inputs which \textit{maximally activate} a set of neurons. In natural imaging, the synthesized inputs show insightful but at the same time hallucinogenic appearing images (for an overview and examples see \cite{olah2017feature}). Simple optimization strategies do not lead to natural looking images. An optimization algorithm does not re-create naturally occurring gradients, or color and intensity similarities in neighboring pixels. Therefore, the outputs of vanilla feature visualization are not insightful and we have tested several methods of regularization. 

First, the network is trained on a data set and its learned weights are saved. During the training, the network has fixed inputs and labels, while the weights of the model are updated until a sufficiently high accuracy is reached. During activation maximization, these roles are changed – the learned weights are fixed, while the input of the model is modified to see what kind of image activates a given channel the most.  Thus, we generate the most salient input \(x^*\) by maximizing the target activation \(f(x)\), while reducing a regularization penalty \(R(x)\):

\[x^* = argmax f(x) - \lambda \cdot R(x)\]

Here \(\lambda\) is a regularization coefficient which we set to 10 throughout this study. The regularization penalty \(R(x)\) calculates the absolute value of the encoded output using L1-regularization. We initialized the optimization with random noise.
We experimented with different regularization techniques and parameters to obtain meaningful feature visualizations. Specifically, we used 5 types of spatial transformations: jitter, random rotation, translation, resize, random resized crop. As well as two types of frequency penalization: switching bilateral filters and total variation denoising. For an overview see Table \ref{tab:reg}. 


\begin{table}
\caption{Overview of the different regularization methods used.}\label{tab:reg}
\begin{tabular}{|l|p{8cm}|}
\hline
\textbf{Name} &  \textbf{Description}\\
\hline
Jitter &  Randomly changes the brightness of the image within a given range.\\
\hline
Rotation & Rotate the image by a random angle within a given range. \\
\hline
Translation & Move the image horizontally or vertically by random value within a given distance.\\
\hline
Resize & Resize the image to a given size.\\
\hline
Random resized crop & Resize the image with a given aspect ratio and crop it randomly within a size range.\\
\hline
Switching bilateral filter (SBF) & A frequency penalization method which smooths the image while, unlike Gaussian blurring, preserving the edges. The SBF, unlike regular bilateral filters, can remove impulse and mixed noise by detecting and replacing only noisy pixels by comparing a pixel with a reference median calculated via its close neighbors \cite{switching}.\\
\hline
Total variation (TV) denoising & A different smoothing technique that preserves the edges of an image by minimizing the total variation across adjacent pixels in an image \cite{vogel1996iterative}. Total variation regularization is best at removing Gaussian noise.\\
\hline
\end{tabular}
\end{table}

\section{Results}
We have generated images using activation maximization on a wide variety of configurations for each regularization method and combinations of them. In the following, we have reduced the number of images to show the main effects we have found in terms of data sets, model hierarchy and regularization techniques. 

\subsection{Data Set Effects and Differences Between Layers}

\begin{figure}
    \centering
    \includegraphics[width=0.85\linewidth]{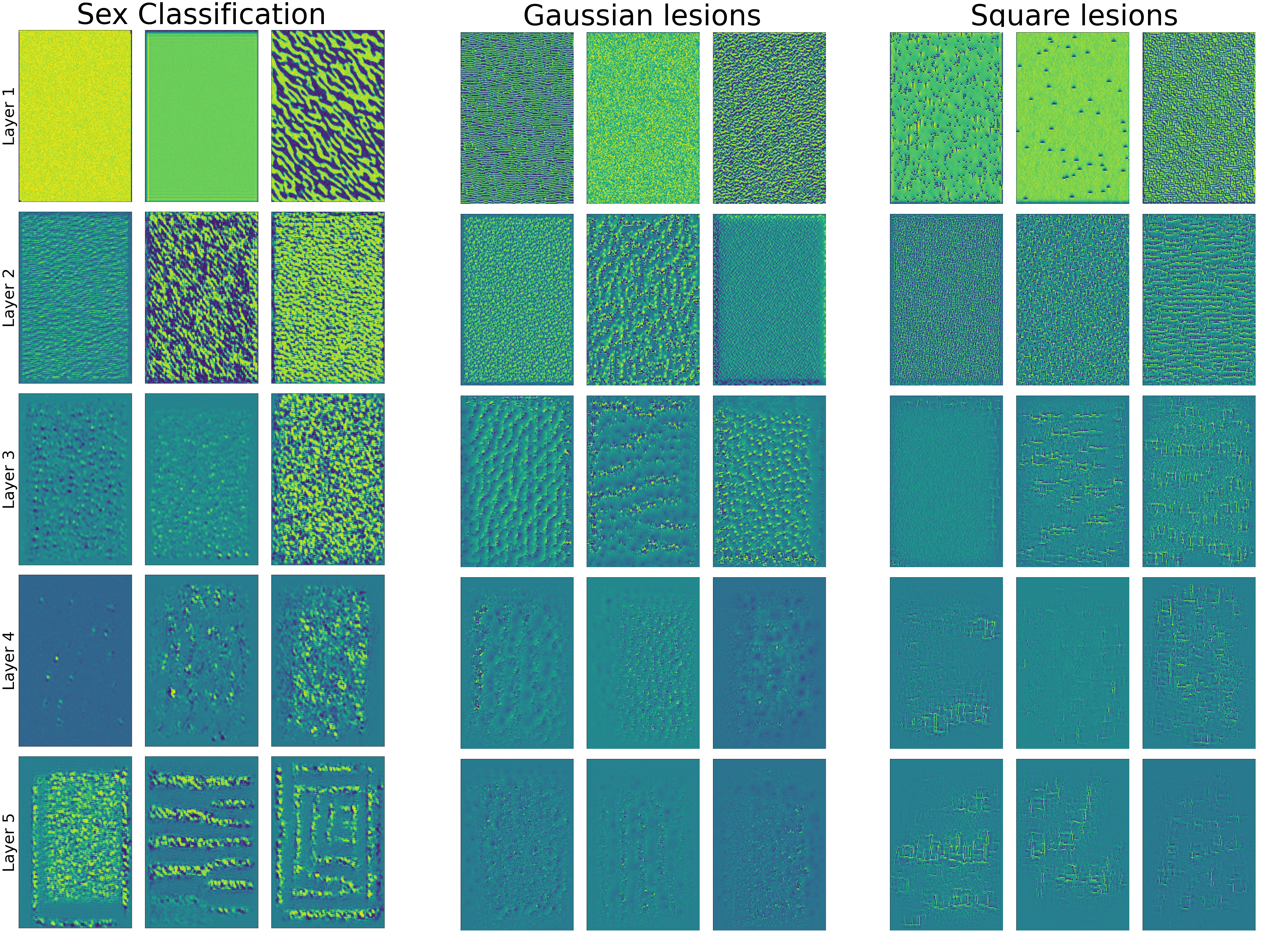}
    \caption{Examples of three random filters of each layer from the sex classification (left), Gaussian lesion (center) and square lesion (right) data sets. The images are generated with 200 iterations each. Figures are best viewed on a large screen.}
    \label{fig:data_sets}
\end{figure}

By comparing the filters across tasks and layer by layer, we can see the effects of the hierarchy of CNNs. As shown in Figure \ref{fig:data_sets}, in the first layer all three models generate patterns of wavy lines, that resemble zebra stripes and are similar to those that invoke neural activity in V1 of the visual cortex. Interestingly, the zebra stripes can have different orientations (also diagonal) and vary in thickness. In addition to that, the Gaussian model generated images with small dots on a neutral background, while the square model generated layers with short straight patterns. Note that sometimes images remain at a nearly uniform value.

In the second layer, the patterns become already seemingly more complex, as this layer allows compositions of first layer features. The images are more granulated and resemble cloth texture. The Gaussian model also generated images with round spots and indents (centre image), while the square model coherently shows square spots and textures (centre image), correctly revealing the shape of the artificial lesions. Considering those spots, it is possible that the second layer detects sharp dips and unevenness in the images and could also contain information about the size of the lesions.

In the third layer, the patterns of the artificial lesions become clearly visible. Especially, in the centre image of the square lesion example, repeating patterns of rectangles in different sizes, but almost square like proportions, populate the whole image. The right image of the Gaussian lesions, on the other hand, shows a dense circular pattern, similar to a honeycomb. We see sharp edges in the square lesion model, likely due to the sharp edges of the lesions in it, while more gradient changes on the sex classification and Gaussian lesion models generate smoother, but more granulated images. Interestingly, the sex classification model generates much blurrier images.

As the layers become more complex, the images they generate also become more intricate and convoluted.
In the fourth layer, the more easily identifiable patterns from the previous layers become ubiquitous, such as the rectangles in the square lesion data set. Some parts of the generated images get blurred out and shapes change from stripes to more delicate features such as circular shapes in the Gaussian data set, while maintaining the same texture. 
Lastly, the fifth layer generates images similar to the fourth layer but with more defined and complex structures. 

Generally, in addition to being able to distinguish the differences between layers, the sharp edges in the square lesion data set is reflected in the generated images, while the gradually disappearing Gaussian lesions create textures with softer edges.

\subsection{Effects of Regularization}

\begin{figure}
    \center
        \begin{turn}{90}
        \includegraphics[width=0.40\linewidth]{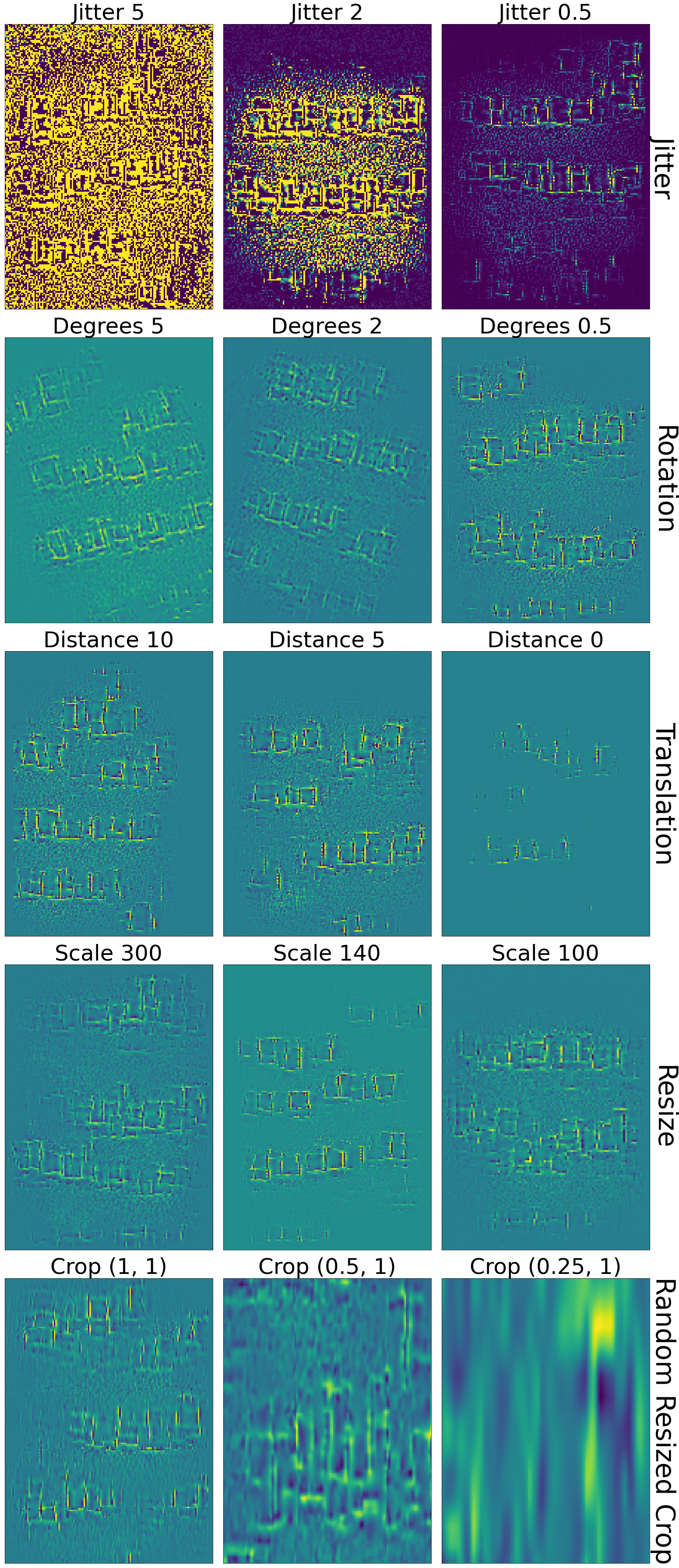}
        \end{turn}
    \caption{The effect of different regularization techniques on the same channel in the fifth layer in the square lesion model.}
    \label{fig:transforms_single}
\end{figure}

\begin{figure}
    \center
        \begin{turn}{90}
        \includegraphics[width=0.40\linewidth]{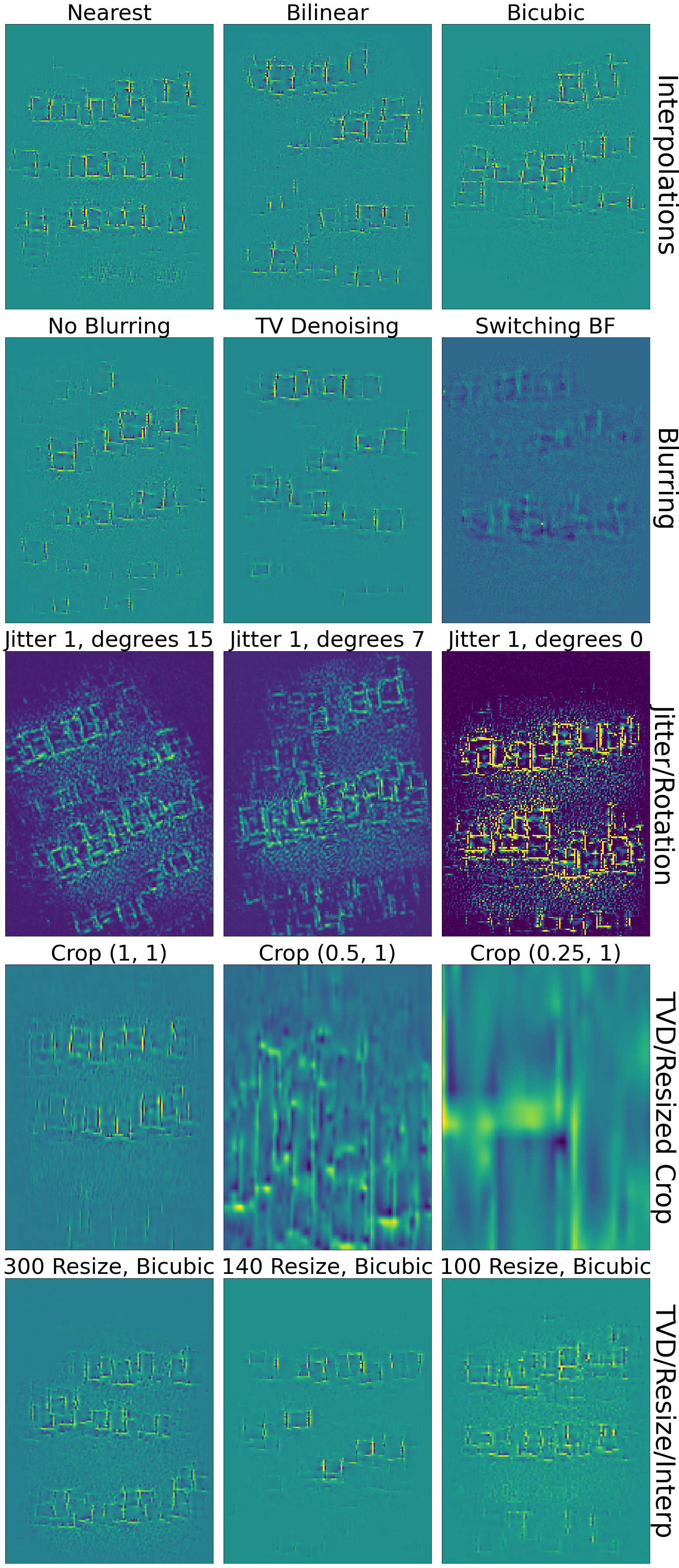}
        \end{turn}
    \caption{The effect of additional regularization techniques and combinations on the same channel in the fifth layer in the square lesion model. TVD stands for total variation denoising.}
    \label{fig:transforms_multiple}
\end{figure}

In Figures \ref{fig:transforms_single} and \ref{fig:transforms_multiple}, the results of several regularization strategies are shown. With the exception of image blurring, all transformations were done at every 50th iteration with 256 iterations in total. As expected, jitter affects the contrast of the resulting images. When jitter brightness is set to 1, the brightness is uniformly adjusted to a random value between 0 and 1. When using a number larger than 1, the value distribution of the final image becomes larger and the whole image becomes brighter. Rotation leaves typical rotation artifacts at the edges of the created images, but also reduces the brightness on the otherwise bright horizontal and vertical edges. Translation also causes artifacts at the edges, but seems to increase the density of the image. Resizing the image also causes edge artifacts but tends to increase the sharpness of the images. Random crops create a strongly zoomed image but also create a smoothing effect. Bilinear interpolation seems to offer more contrast than bicubic, while still staying fine-grained, unlike nearest interpolation. Both frequency penalization methods can remove Gaussian, impulse and mixed noise. It appears that TV denoising smoothed the image more subtly and its effects are easier to notice in the high contrast areas. On the other hand, SBF smoothes the image too drastically, possibly due to low pixel values in the background. 

Lastly, we show combinations of methods in the three right-most columns of Figure \ref{fig:transforms_multiple}. Interestingly, jitter seems to smooth over some of the edge artifacts from the higher amount of rotation. On the other hand, rotation reduces the sharp brightness effects from the jitter. Combining resizing with bicubic interpolation leads to smoother results than resizing by itself.


\section{Discussion}
In the ML-based analysis of neuroimaging data, feature visualization has the potential to improve the detection of disease defining patterns and thus to increase the interpretability of ML models. In both experiments with artificial lesions, the synthesized images were clearly referencing the lesions and respecting their shape, even in the case of Gaussian lesions which are harder to detect for a human. The synthesized images of lower level neurons in our study are highly similar to those studies using natural images \cite{olah2017feature} which supports the general application of CNNs on MRI data. It can also be an indicator for the success of transfer learning across domains. The difference in dimensionality between the studied data sets seems to be negligible and feature visualization is easily convertible to 3D.

Furthermore, the generated images can be an additional method of validation. Although, all models have achieved high accuracies, the generated patterns on the artificial lesion classifications are much sharper than on the sex classification. This might indicate that the model for sex classification has not yet saturated as much as the other two models, due to a much smaller data set while having a more challenging task. Experiments in \cite{schulz2020different} have shown that different algorithms do not saturate on sex classification even with 8,000 samples. Similar to how CNN filter values become smoother with better performance, we would expect that generated images from feature visualization show a similar behavior. Moreover, the generated images with almost uniform distributions are a potential sign of overparameterization of the model. By analysing a model's feature visualizations, data scientists can validate whether the models are suitable for clinical practice. However, at the moment higher level patterns for the sex classification task are difficult to interpret. Combinations of different methods of regularization might reveal interpretable patterns in further studies, in the same way certain combinations revealed interpretable patters in natural images. Potential directions for future studies include training on larger data sets of clinical tasks, as well as assessment of further regularization methods. In addition, future studies should involve scenarios where neuropathological indicators can be clearly described such as the hummingbird sign in progressive supranuclear palsy.

\bibliographystyle{splncs04}
\bibliography{main}

\begin{thebibliography}{10}
\providecommand{\url}[1]{\texttt{#1}}
\providecommand{\urlprefix}{URL }
\providecommand{\doi}[1]{https://doi.org/#1}

\bibitem{ALFAROALMAGRO2018400}
Alfaro-Almagro, F., Jenkinson, M., Bangerter, N.K., Andersson, J.L., Griffanti,
  L., Douaud, G., Sotiropoulos, S.N., Jbabdi, S., Hernandez-Fernandez, M.,
  Vallee, E., Vidaurre, D., Webster, M., McCarthy, P., Rorden, C., Daducci, A.,
  Alexander, D.C., Zhang, H., Dragonu, I., Matthews, P.M., Miller, K.L., Smith,
  S.M.: {Image Processing and Quality Control for the First 10,000 Brain
  Imaging Datasets from UK Biobank}. NeuroImage  \textbf{166},  400--424
  (2018). \doi{https://doi.org/10.1016/j.neuroimage.2017.10.034},
  \url{https://www.sciencedirect.com/science/article/pii/S1053811917308613}

\bibitem{10.3389/fnagi.2019.00194}
Böhle, M., Eitel, F., Weygandt, M., Ritter, K.: {Layer-Wise Relevance
  Propagation for Explaining Deep Neural Network Decisions in MRI-Based
  Alzheimer's Disease Classification}. Frontiers in Aging Neuroscience
  \textbf{11}, ~194 (2019). \doi{10.3389/fnagi.2019.00194},
  \url{https://www.frontiersin.org/article/10.3389/fnagi.2019.00194}

\bibitem{dyrba2021improving}
Dyrba, M., Hanzig, M., Altenstein, S., Bader, S., Ballarini, T., Brosseron, F.,
  Buerger, K., Cantré, D., Dechent, P., Dobisch, L., Düzel, E., Ewers, M.,
  Fliessbach, K., Glanz, W., Haynes, J.D., Heneka, M.T., Janowitz, D., Keles,
  D.B., Kilimann, I., Laske, C., Maier, F., Metzger, C.D., Munk, M.H.,
  Perneczky, R., Peters, O., Preis, L., Priller, J., Rauchmann, B., Roy, N.,
  Scheffler, K., Schneider, A., Schott, B.H., Spottke, A., Spruth, E.J., Weber,
  M.A., Ertl-Wagner, B., Wagner, M., Wiltfang, J., Jessen, F., Teipel, S.J.:
  {Improving 3D convolutional Neural Network Comprehensibility via Interactive
  Visualization of Relevance Maps: Evaluation in Alzheimer's Disease} (2021)

\bibitem{EITEL2019102003}
Eitel, F., Soehler, E., Bellmann-Strobl, J., Brandt, A.U., Ruprecht, K., Giess,
  R.M., Kuchling, J., Asseyer, S., Weygandt, M., Haynes, J.D., Scheel, M.,
  Paul, F., Ritter, K.: {Uncovering Convolutional Neural Network Decisions for
  Diagnosing Multiple Sclerosis on Conventional MRI Using Layer-Wise Relevance
  Propagation}. NeuroImage: Clinical  \textbf{24},  102003 (2019).
  \doi{https://doi.org/10.1016/j.nicl.2019.102003},
  \url{https://www.sciencedirect.com/science/article/pii/S2213158219303535}

\bibitem{goodfellow2015explaining}
Goodfellow, I.J., Shlens, J., Szegedy, C.: {Explaining and Harnessing
  Adversarial Examples} (2015)

\bibitem{graber2009teaching}
Graber, J.J., Staudinger, R.: {Teaching Neuroimages:“Penguin” or
  “hummingbird” Sign and Midbrain Atrophy in Progressive Supranuclear
  Palsy}. Neurology  \textbf{72}(17),  e81--e81 (2009)

\bibitem{kindermans2017learning}
Kindermans, P.J., Schütt, K.T., Alber, M., Müller, K.R., Erhan, D., Kim, B.,
  Dähne, S.: {Learning How to Explain Neural Networks: PatternNet and
  PatternAttribution} (2017)

\bibitem{switching}
Lin, C.H., Tsai, J.S., Chiu, C.T.: {Switching Bilateral Filter with a
  Texture/Noise Detector for Universal Noise Removal}. In: 2010 IEEE
  International Conference on Acoustics, Speech and Signal Processing. pp.
  1434--1437 (2010). \doi{10.1109/ICASSP.2010.5495475}

\bibitem{lundberg2017unified}
Lundberg, S., Lee, S.I.: {A Unified Approach to Interpreting Model Predictions}
  (2017)

\bibitem{mahendran2015understanding}
Mahendran, A., Vedaldi, A.: {Understanding Deep Image Representations by
  Inverting Them}. In: Proceedings of the IEEE conference on computer vision
  and pattern recognition. pp. 5188--5196 (2015)

\bibitem{montavon2019layer}
Montavon, G., Binder, A., Lapuschkin, S., Samek, W., M{\"u}ller, K.R.:
  {Layer-Wise Relevance Propagation: an Overview}. Explainable AI:
  interpreting, explaining and visualizing deep learning pp. 193--209 (2019)

\bibitem{mordvintsev2015inceptionism}
Mordvintsev, A., Olah, C., Tyka, M.: {Inceptionism: Going Deeper into Neural
  Networks} (2015)

\bibitem{olah2017feature}
Olah, C., Mordvintsev, A., Schubert, L.: {Feature Visualization}. Distill
  \textbf{2}(11), ~e7 (2017)

\bibitem{schulz2020different}
Schulz, M.A., Yeo, B.T., Vogelstein, J.T., Mourao-Miranada, J., Kather, J.N.,
  Kording, K., Richards, B., Bzdok, D.: {Different Scaling of Linear Models and
  Deep Learning in UKBiobank Brain Images Versus Machine-Learning Datasets}.
  Nature communications  \textbf{11}(1),  1--15 (2020)

\bibitem{DBLP:journals/corr/ShrikumarGK17}
Shrikumar, A., Greenside, P., Kundaje, A.: {Learning Important Features Through
  Propagating Activation Differences}. CoRR  \textbf{abs/1704.02685} (2017),
  \url{http://arxiv.org/abs/1704.02685}

\bibitem{simonyanzisserman2015}
Simonyan, K., Zisserman, A.: {Very Deep Convolutional Networks for Large-Scale
  Image Recognition}. 3rd International Conference on Learning Representations
  (ICLR) pp. 1--14 (2015), \url{https://arxiv.org/pdf/1409.1556.pdf}

\bibitem{tyka2016bilateral}
Tyka, M.: {Class Visualization with Bilateral Filters}.
  \url{https://mtyka.github.io/deepdream/2016/02/05/bilateral-class-vis.html},
  accessed: 2021-06-23

\bibitem{vogel1996iterative}
Vogel, C.R., Oman, M.E.: {Iterative Methods for Total Variation Denoising}.
  SIAM Journal on Scientific Computing  \textbf{17}(1),  227--238 (1996)

\end{thebibliography}
\end{document}